\documentclass[10pt,twocolumn,letterpaper]{article}
\pdfoutput=1

\usepackage[pagenumbers]{cvpr} 

\usepackage{graphicx}
\usepackage{amsmath}
\usepackage{amssymb}
\usepackage{booktabs}

\usepackage{graphicx, amsmath, amssymb, caption, multirow, overpic, textpos}
\usepackage[table]{xcolor}
\usepackage[british, american]{babel}
\definecolor{citecolor}{RGB}{119,185,0}
\definecolor{linkcolor}{HTML}{ED1C24}
\usepackage[pagebackref=false, breaklinks=true, colorlinks,
            citecolor=citecolor, linkcolor=linkcolor, bookmarks=false]{hyperref}

\newlength\savewidth\newcommand\shline{\noalign{\global\savewidth\arrayrulewidth
  \global\arrayrulewidth 1pt}\hline\noalign{\global\arrayrulewidth\savewidth}}
\newcommand{\tablestyle}[2]{\setlength{\tabcolsep}{#1}\renewcommand{\arraystretch}{#2}\centering\footnotesize}
\renewcommand{\paragraph}[1]{\vspace{1.25mm}\noindent\textbf{#1}}

\newcolumntype{x}[1]{>{\centering\arraybackslash}p{#1pt}}
\newcolumntype{y}[1]{>{\raggedright\arraybackslash}p{#1pt}}
\newcolumntype{z}[1]{>{\raggedleft\arraybackslash}p{#1pt}}

\usepackage[capitalize]{cleveref}
\crefname{section}{Sec.}{Secs.}
\Crefname{section}{Section}{Sections}
\Crefname{table}{Table}{Tables}
\crefname{table}{Tab.}{Tabs.}

\begin{document}

\title{1st Place Solution of The Robust Vision Challenge 2022 \\
Semantic Segmentation Track}

\author
{
Junfei Xiao${^{1}}$
\quad
Zhichao Xu${^{3}}$
\quad
Shiyi Lan${^{2}}$
\quad
Zhiding Yu${^{2}}$
\quad
Alan Yuille${^{1}}$
\quad
Anima Anandkumar${^{2}}$
\\$^{1}$Johns Hopkins University \qquad ~$^{2}$NVIDIA \qquad ~$^{3}$Fudan University
}
\maketitle

\begin{abstract}
   This report describes the winning solution to the Robust Vision Challenge (RVC) semantic segmentation track at ECCV 2022. Our method adopts the FAN-B-Hybrid model as the encoder and uses SegFormer as the segmentation framework. The model is trained on a composite dataset consisting of images from 9 datasets (ADE20K, Cityscapes, Mapillary Vistas, ScanNet, VIPER, WildDash 2, IDD, BDD, and COCO) with a simple dataset balancing strategy. All the original labels are projected to a 256-class unified label space, and the model is trained using a cross-entropy loss. Without significant hyperparameter tuning or any specific loss weighting, our solution ranks the first place on all the testing semantic segmentation benchmarks from multiple domains (ADE20K, Cityscapes, Mapillary Vistas, ScanNet, VIPER, and WildDash 2). The proposed method can serve as a strong baseline for the multi-domain segmentation task and benefit future works. Code will be available at \url{https://github.com/lambert-x/RVC_Segmentation}.
\end{abstract}

\section{Introduction}
\label{sec:intro}

In the past few years, advances in deep learning have led to significant progress in visual recognition. However, the robustness of state-of-the-art deep learning models remains an open issue. On the one hand, real-world applications require models to be deployed ``in the wild''. On the other hand, many current deep models have been shown to be brittle to distributional shifts and natural perturbations. This phenomenon raised considerable interest in open problems such as domain generalization and adaptation.

There is rich literature in domain generalization~\cite{wang2022generalizing,zhou2022domain} where popular methods include, but are not limited to: domain randomization, domain invariant representation learning, disentanglement learning and meta learning, etc. One approach related to this work is multi-dataset training~\cite{lambert2020mseg}, in which the authors show that a simple combination of multiple datasets with label space alignment can outperform strong domain generalization approaches.

Another interesting trend is the recent surge of Vision Transformers (ViTs). Several works~\cite{paul2022vision,naseer2021intriguing,bai2021transformers,xie2021segformer} almost simultaneously pointed out that ViTs demonstrate surprisingly strong robustness to out-of-distribution scenarios. For example, SegFormer~\cite{xie2021segformer} demonstrates significantly better results over CNN-based strong methods in Cityscapes-C, a more challenging variant of Cityscapes contaminated by 16 types of natural corruption. More recently, \cite{zhou2022understanding} introduced the fully attentional network (FAN), a family of ViT backbones with state-of-the-art accuracy and robustness in both image classification and downstream tasks.

This report describes the winning solution to the RVC 2022 semantic segmentation track. This year, the challenge features benchmarking of a single semantic segmentation model on six datasets, spanning both indoor/outdoor and synthetic/real. Thus, it presents a great challenge to the generalization capability of a model over different domains. Our solution is inspired by the above advances in both multi-dataset training and ViTs, as will be detailed in the rest of the report.

\section{Method}
\label{sec:method}

\begin{table}[t]
\centering
\small
\begin{tabular}{llcc}
\multirow{2}{*}{Dataset} & \multirow{2}{*}{Scenes} 
  & {\#Images }
  & {\#Class}

  \\
  & &  (train/val) & (orig-proj) \\
\hline
COCO~\cite{caesar2018coco,lin2014microsoft}       & Natural   &  118287/5000
  & 201$\rightarrow$133\\
ADE20K~\cite{zhou2017scene}       & Natural   &  20210/2000
  & 151$\rightarrow$146 \\
Cityscapes~\cite{cordts2016cityscapes}   & Driving  &   2975/500 
  & 34$\rightarrow$31    \\
Vistas~\cite{neuhold2017mapillary}       & Driving  &  18000/2000 
  &  66$\rightarrow$64 \\
BDD~\cite{yu2020bdd100k}   & Driving  &   7000/1000 
  & 19$\rightarrow$19     \\
  
IDD~\cite{varma2019idd}   & Driving  &   6993/981 
  & 39$\rightarrow$26    \\
WildDash 2~\cite{zendel2018wilddash}     & Driving  &   3413/857 
  & 34$\rightarrow$31     \\

ScanNet~\cite{dai2017scannet}      & Indoor &  19466/5436 
  & 41$\rightarrow$41    \\
VIPER~\cite{richter2017playing}        & Artificial &  13367/4959 
  & 32$\rightarrow$32    \\

\end{tabular}
\caption{\textbf{Datasets overview.}
  A total of 9 datasets across natural, driving, indoor, and artificial scenes are used for training and validating the model.
  Class count denotes the number of classes in the original label space
  and the projected label space.
  }
\label{tab:datasets}
\end{table}

\begin{table*}[!t]
\centering

 \tablestyle{8pt}{1.05}
\begin{tabular}{@{}lcccccccc@{}}
\toprule
  & \multicolumn{6}{c}{ \textsc{RVC Test Datasets} } \\
\textsc{Method Name} &\textsc{Year/Rank}   &  \textsc{ADE20K} & \textsc{Cityscapes}   & \textsc{Mapillary} & \textsc{ScanNet} & \textsc{VIPER} & \textsc{WildDash 2}  \\
\midrule

\textsc{MSeg1080\_RVC} \cite{lambert2020mseg} & {2020 / 2nd} & {{33.18}}   & {{80.7}}  & {34.19}   & {48.5}   & {40.7}  & {34.71} \\
\textsc{SN\_RN152pyrx8\_RVC} \cite{bevandic2020multi} & {2020 / 1st} & {31.12}   & {74.7}  & {{40.43}}     & {{54.6}}      & {{62.5}}  & {{42.29}}  \\

\textsc{FAN\_NV\_RVC} (Ours) & {2022 / 1st} & \textbf{43.46}   & \textbf{82.0}     & \textbf{{55.27}}     & \textbf{{58.6}}      & \textbf{{69.8}}  & \textbf{{47.5}}  \\

\bottomrule
\end{tabular}
 \caption{\textbf{Comparison with previous methods.}  Measured by class mIoU. The best number in each column is highlighted in bold.} 
\label{tab:results}
\end{table*}

\paragraph{Backbone.}
We adopt FAN-B-Hybrid~\cite{zhou2022understanding} as our backbone encoder due to its great robustness on multiple benckmarks (ImageNet-C~\cite{hendrycks2018benchmarking}, Cityscapes-C, etc.). The backbone is initialized with the weight pretrained on ImageNet-22K and fine-tuned on ImageNet-1K (the checkpoint is provided in the official github repository\footnote{\url{https://github.com/NVlabs/FAN}}).

\paragraph{Segmentation framework.}
We use SegFormer~\cite{xie2021segformer} as the segmentation framework. It uses simple but effective multilayer perceptron (MLP) decoders to fuse multi-level features (the outputs of the early Convolution blocks, last FAN Transformer block and the final class attention block output) and predict the semantic segmentation mask. The reader may refer to the official github of FAN (segmentation folder) for more details. Cross-entropy loss is used for training the model.

\paragraph{Training set.}
The model is trained on a composite dataset with all the training images from ADE20K, Cityscapes, Mapillary Vistas, ScanNet, VIPER, WildDash 2, IDD, BDD, and COCO. An overview of all the datasets involved is shown in \Cref{tab:datasets}. The datasets vary largely in size (COCO is more than 30 times larger than WildDash 2). To alleviate the imbalanced data size issue for better model generalization, we adopt a simple dataset resizing strategy - repeat each dataset $(120,000// len(dataset))$ times.

\paragraph{Unified label space.}
We directly use the unified label space provided in the official RVC github repository\footnote{\url{https://github.com/ozendelait/rvc_devkit}} (with some minor corrections), which has 256 classes. This label space is naive and noisy for relabeling some fine-grained classes in their original label space. 

\paragraph{Post-processing.} All predicted segmentation maps are projected from the unified label space to the original label space of every dataset.

\section{Implementation Details}
\label{sec:imple_details}
We built our codebase with MMSegmentation~\cite{contributors2020mmsegmentation}. The length of the training process is 80,000 iterations, while the first half training is without BDD and IDD datasets. \Cref{tab:imple_details} provides detailed information about the optimizer and hyperparameter settings. Training and testing data augmentations are detailed in \Cref{tab:train_aug} and \Cref{tab:test_aug}. The model is trained on 64 V100 GPUs (32G), and the whole training procedure takes $\sim$35 hours.

\begin{table}[ht]
\tablestyle{6pt}{1.02}
\small
\begin{tabular}{y{96}|y{68}}
Config & Setting \\
\shline
Optimizer & AdamW \cite{loshchilov2018decoupled} \\
Learning rate & 6e-5 \\
Weight decay & 0.01 \\
Optimizer momentum & $\beta_1, \beta_2{=}0.9, 0.999$ \\
Batch size & 64 \\
Learning rate schedule & Poly \cite{chen2017deeplab}\\
Warmup iters \cite{goyal2017accurate} & 1500 \\
\end{tabular}
\vspace{-.5em}
\caption{\textbf{Optimizer \& hyper-parameters details.}}
\label{tab:imple_details} \vspace{-.5em}
\end{table}

\begin{table}[ht]
\tablestyle{8pt}{1.05}
\small
\begin{tabular}{c|c}
 Operation & Setting \\ 
\shline
Resize &  Scale: (2048, 1024), Ratio: (0.5, 2.0) \\
RandomCrop &  Crop size: (1024, 1024) \\
RandomFlip & Prob: 0.5 \\
PhotoMetricDistortion & Default
\end{tabular}
\caption{\textbf{Training data augmentations.}}
\label{tab:train_aug} 
\vspace{-.5em}
\end{table}

\begin{table}[ht]
\tablestyle{12pt}{1.2}
\small
\begin{tabular}{c|c}
 Operation & Setting \\ 
\shline
 Resize & Scale: (2048, 1024) \\
 Multi-scale &  Ratios: (0.5, 0.75, 1.0, 1.25, 1.5, 1.75) \\
 Flip & True
\end{tabular}
\caption{\textbf{Testing data augmentations.}}
\label{tab:test_aug} 
\vspace{-.5em}
\end{table}

\section{Results}
\label{sec:results}

We compare our method with the winning solutions of RVC 2020 in \Cref{tab:results}. Our method makes solid improvements on all six benchmarks and beats them all by a large margin.

\section{Conclusion}
\label{sec:conclusion}

In this report, we described the winning solution of the RVC 2022 Semantic Segmentation Track. Our result shows that Vision Transformer models (in our case, FAN), when coupled with multi-dataset training, exhibit strong robustness and generalization at scales in semantic segmentation. Our work again echoes recent discoveries of improved robustness and representation in ViTs. However, it is worth noting that the training computation and memory consumption have become important challenges as the data and label space become large. The efficiency on devices during deployment also presents another challenge for real-world applications of current ViT models.

{\small
\bibliographystyle{ieee_fullname}
\bibliography{ref}
}

\end{document}